\documentclass[10pt, a4paper]{article}
\usepackage{lrec2022} 
\usepackage{multibib}
\newcites{languageresource}{Language Resources}
\usepackage{graphicx}
\usepackage{tabularx}
\usepackage{soul}
\usepackage{titlesec}
\titleformat{\section}{\normalfont\large\bfseries\center}{\thesection.}{1em}{}
\titleformat{\subsection}{\normalfont\SmallTitleFont\bfseries\raggedright}{\thesubsection.}{1em}{}
\titleformat{\subsubsection}{\normalfont\normalsize\bfseries\raggedright}{\thesubsubsection.}{1em}{}
\renewcommand\thesection{\arabic{section}}
\renewcommand\thesubsection{\thesection.\arabic{subsection}}
\renewcommand\thesubsubsection{\thesubsection.\arabic{subsubsection}}

\usepackage{epstopdf}
\usepackage[utf8]{inputenc}

\usepackage{hyperref}
\usepackage{xstring}
\usepackage{times}
\usepackage{latexsym}

\usepackage{kotex}
\usepackage{amsmath}
\usepackage{amssymb}
\usepackage{bbm}
\usepackage{bm} 
\usepackage{graphicx}
\DeclareMathOperator*{\argmax}{argmax}

\usepackage{multirow}
\usepackage{paralist}
 \usepackage {booktabs}
\usepackage[normalem]{ulem}
\usepackage{subcaption}

\usepackage{color}

\title{Learning How to Translate North Korean through South Korean}

\name{Hwichan Kim$^	\dagger$, Sangwhan Moon$^{\ddag,\ast}$, Naoaki Okazaki$^\ddag$, Mamoru Komachi$^	\dagger$} 

\address{$\dagger$Tokyo Metropolitan University, 6-6 Asahigaoka, Hino, Tokyo 191-0065, Japan\\
$\ddag$Tokyo Institute of Technology, 2-12-1 Ookayama, Meguro, Tokyo 152-8550, Japan \\
$\ast$Google LLC, 1600 Amphitheatre Parkway Mountain View, CA 94043, USA \\
        kim-hwichan@ed.tmu.ac.jp, sangwhan@iki.fi, okazaki@c.titech.ac.jp, komachi@tmu.ac.jp\\}

\abstract{
South and North Korea both use the Korean language. However, Korean NLP research has focused on South Korean only, and existing NLP systems of the Korean language, such as neural machine translation (NMT) models, cannot properly handle North Korean inputs. Training a model using North Korean data is the most straightforward approach to solving this problem, but there is insufficient data  to train NMT models. 
In this study, we create data for North Korean NMT models using a comparable corpus.
First, we manually create evaluation data for automatic alignment and machine translation. Then, we investigate automatic alignment methods suitable for North Korean. Finally, we verify that a model trained by North Korean bilingual data without human annotation can significantly boost North Korean translation accuracy compared to existing South Korean models in zero-shot settings.
 \\ \newline  }

\begin{document}

\maketitleabstract

\section{Introduction}
South and North Koreans use the same Korean language, with the same grammar.
However, there are some differences between South and North Korean vocabularies and spelling rules \cite{Lee1990,YUN2019100918}.

Several NLP researchers have recently been working on the Korean language. For example, the Workshop of Asian Translation \cite{nakazawa-etal-2021-overview} has been conducting a series of shared tasks annually, including on Korean language variants.
However, these studies are focused exclusively on the South Korean language; none of the developed NLP systems support the North Korean language. For example, public neural machine translation (NMT) systems cannot translate North Korean-specific words (Table \ref{translation}).
Although training models on North Korean data is a simple and effective way to improve the quality of North Korean translation, parallel data for training are unavailable\footnote{\newcite{kim-etal-2020-zero} proposed a North Korean and English evaluation dataset for machine translation by manually rewriting sentences of a South Korean dataset to conform with North Korean spelling rules. However, as they are written from a South Korean dataset, the sentences in the data are not considered of North Korean provenance.}.

In this study, we tackle North Korean to English and Japanese bilingual data creation from comparable corpora to train a North Korean NMT model. 
Our contribution in this paper is threefold: 
\begin{inparaenum}[(1)]
\item We manually create North Korean evaluation data for the development of MT systems.
\item We investigate automatic article and sentence alignment methods suitable for North Korean, and create a small amount of North Korean parallel training data using a method that achieved the highest alignment quality.
\item We compare North Korean to English and Japanese NMT models and show that our North Korean data can significantly enhance the translation quality when used in conjunction with South Korean datasets. 
\end{inparaenum}

\begin{table*}[tp]
\begin{center}
\scalebox{1.0}{
\begin{tabular}{ll}
\toprule
NK source & 4월 24일 \textbf{로씨야}련방 \textbf{울라지보스또크}시에 도착하시였다.  \\
Reference & He arrived at \textbf{Vladivostok}, the \textbf{Russian} Federation on Wednesday.\\\midrule
SK & He arrived at the city of \uwave{Ulazibosto} on April 24th. \\
SK$\rightarrow$NK  & He  arrived in \textbf{Vladivostok},  the \textbf{Russian} Federation  on April 24. \\
Google & On April 24th, you arrived in the city of \uwave{Ulagivostok} in the \textbf{Russian} Federation. \\
NAVER & On April 24th, he arrived at \uwave{Ulajibos Tok} City, a \uwave{training room for RoC}. \\
\bottomrule
\end{tabular}
}
\end{center}
\caption{
Translation example. SK denotes the South Korean model, and SK$\rightarrow$NK  denotes the model fine-tuned by our North Korean data. The squiggles indicate mistranslated words.
}
\label{translation}
\end{table*}

\section{Related Work}
\subsection{Automatic Parallel Corpus Alignment}
Building NMT systems requires parallel data consisting of parallel sentences. However, the manual creation of parallel sentences is costly and time-consuming. Consequently, research on the automatic alignment of parallel sentences from parallel documents is actively underway.  
The typical methods proposed to date are based on using a bilingual dictionary for sentence alignment \cite{chen-1993-aligning,etchegoyhen-azpeitia-2016-set,azpeitia-etal-2017-weighted}. 
These methods translate source words to target words using a bilingual dictionary, and then align the sentences based on the similarity between the translated sentences.
\newcite{sennrich-volk-2010-mt}, \newcite{gomes-lopes-2016-first}, and \newcite{karimi-etal-2018-extracting} used an existing machine translation (MT) system instead of a bilingual dictionary.
If we adopt this approach to North Korean alignment, using a South Korean MT system is a possible approach because there are no publicly available North Korean MT systems or models.

The alignment methods based on cross-lingual representations are useful methods that map sentences to the cross-lingual semantic space and align them according to their closeness \cite{schwenk-douze-2017-learning,schwenk-2018-filtering,10.1162/tacl_a_00288,sun-etal-2021-parallel}.
One such alignment approach using language-agnostic sentence representations (LASER) \cite{10.1162/tacl_a_00288} achieved state-of-the-art performance in the building and using comparable corpora (BUCC) task \cite{zweigenbaum-etal-2017-overview}.\footnote{A shared task on parallel sentence extraction from parallel documents.} 
This approach used representations of a multilingual NMT model encoder as the cross-lingual representations.
In this study, we compare these two approaches of using an MT system and LASER, and create North Korean training parallel data through an approach that achieved the highest alignment quality.

\subsection{Machine Translation for Dialects}
In addition to South and North Korean, several languages have dialects such as  Brazilian and European Portuguese, Canadian and European French.
\newcite{lakew-etal-2018-neural} demonstrated that the translation accuracy is dropped when using the different dialect's training data with target one. 

One of the reasons for this problem is the spelling, lexical, and grammar divergence between dialects.
Therefore, to mitigate the reduction in translation accuracy, the differences between the dialects must be absorbed. 
Rule-based transformation between dialects is one of the approaches for achieving this \cite{marujo-etal-2011-bp2ep,6473708}. 
Additionally, several studies have attempted to construct an MT system between the dialects \cite{durrani-etal-2010-hindi,popovic-etal-2016-language,HARRAT2019262}.
However, rule-based transformation cannot address the differences between the vocabularies, and to construct a machine translation system, a parallel corpus is necessary between the dialects. 

Transfer learning is also a useful approach if there are parallel data between the dialect and target language.
Transfer learning, which is an approach to fine-tune the NMT model trained by the parallel corpus of another language pair (transfer source) with the one of low-resource-language pair (transfer destination), is an effective approach for improving the accuracy in a low-resource-language scenario.  
Previous studies have demonstrated that transfer learning works efficiently when the transfer source and destination languages are linguistically similar \cite{zoph-etal-2016-transfer,dabre-etal-2017-empirical}. 
Dialects typically have almost the same grammar and many vocabularies in common.
In fact, \newcite{lakew-etal-2018-neural} showed that the transfer learning is effective for the dialects of Portuguese and French.

Since the differences between South and North Korean languages are not only in the grammar but also vocabulary, it is difficult to absorb the differences with only the rule-based transformation.
Furthermore, there is no available bilingual dictionary and parallel data between the South and North Korean.  
However, we can construct parallel data between North Korean and a target language using North Korean news articles.
Consequently, in this study, we adopt the transfer learning approach, using South Korean and the target language NMT model as the transfer source.

\section{North Korean Parallel Corpus Construction}
In this study, we create North Korean parallel corpus from North Korean news articles.
We use a news portal, Uriminzokkiri\footnote{\url{http://www.uriminzokkiri.com/}}, that publishes news articles from various North Korean (NK) news sources\footnote{In this study, we used articles from September 2017 to June 2021, when we started our experiment. 
The URLs of articles prior to September 2017 are available, but we are unable to access them. 
We obtained permission to re-distribute the article data.
}. These articles are translated into English (EN), Chinese, Russian and Japanese (JA). 
In this study, we use North Korean, English, and Japanese articles.

\begin{table}[tp]
\begin{center}
\begin{tabular}{lcc}
\toprule
Language & Articles & Sentences  \\\midrule
North Korean & 408 & 6,622 \\
English & 414 &  6,770 \\
Japanese &415 & 6,220 \\
\bottomrule
\end{tabular}
\end{center}
\caption{
Number of articles and sentences. The number of articles differs because unique articles exist in each language.
}
\label{article_statics}
\end{table}

Table \ref{article_statics} lists the total numbers of articles and their sentences.
One of the problems with the data sourced from this site is that articles and sentences are not aligned between North Korean and each of the other languages. 
Therefore, we manually and automatically align them to create North Korean parallel corpus. 

\subsection{Manual Evaluation Data Alignment}
We manually align the NK--EN and NK--JA articles and sentences to create evaluation data for MT. At first, we align all articles (Table \ref{article_statics}). Then, we randomly sample the sentences from the English and Japanese articles and manually select the parallel sentences from the North Korean articles. The alignments between these languages require an annotator that can read and understand Korean, English, and Japanese. 
Therefore, we assign a trilingual annotator---a Korean living in Japan who is enrolled in a computer science master's program. To measure inter-annotator agreement, we additionally ask two bilingual Koreans to perform annotations\footnote{One of the Koreans is a native Japanese speaker and the other has worked in the United States of America.}. 

Furthermore, we create evaluation data for North Korean article and sentence alignments to investigate automatic alignment methods suitable for North Korean.
For the article alignment evaluation, we use the aligned articles.
For the sentence alignment evaluation, we select one article and manually align sentences it contained.
We ask this alignment to the trilingual annotator.

\begin{table}
\begin{center}
\scalebox{0.9}{
\begin{tabular}{lccc|ccc}
\toprule
&  \multicolumn{2}{c}{Mono} & Para &  \multicolumn{2}{c}{Mono} & Para \\
  &  NK & EN & NK--EN &  NK & JA  & NK--JA \\\midrule 
Sentences & 290 & 300 & 285  & 143 & 100 & 100 \\
Articles & 408 & 414  & 359  & 408 & 415  & 356 \\
\bottomrule
\end{tabular}
}
\end{center}
\caption{Manually created sentence and article alignment evaluation data. These figures indicate the numbers of monolingual sentences and articles (Mono) in each language and annotated alignments of parallel sentences and articles (Para).}
\label{manual_alignment}
\end{table}

\begin{table}
\begin{center}
\scalebox{1.0}{
\begin{tabular}{lcc|cc}
\toprule
  & \multicolumn{2}{c}{NK--EN} &  \multicolumn{2}{|c}{NK--JA} \\\midrule
 & sentence & article & sentence & article \\
Naive  & 0.1 & -  & 0.5 & -\\
LASER & 96.7 & 85.3 & 96.9 & 96.7 \\
to-SK & 94.3  &  96.9 & 97.4 & 98.2\\
from-SK & 96.1 & 95.7 &  95.3 & 96.3 \\
bidi-SK & \textbf{96.9} & \textbf{97.6}${}^\text{\dag}$ & \textbf{97.5}  & \textbf{98.4}${}^\text{\dag}$ \\
\bottomrule
\end{tabular}
}
\end{center}
\caption{Sentence and article alignment F1 scores.  
 $\text{\dag}$ indicates the statistical significance ($p$ $<$ 0.05) between the bidi-SK and LASER. 
}
\label{sentence_alignment_acc2}
\end{table}

\subsection{Automatic Training Data Alignment through South Korean NMT}

Previous studies have proposed several alignment approaches, such as using a bilingual dictionary \cite{chen-1993-aligning,azpeitia-etal-2017-weighted}. Furthermore, approaches have been proposed that use the representations from the cross-lingual models, which are trained with a supervised method \cite{schwenk-2018-filtering,10.1162/tacl_a_00288} or an unsupervised method \cite{keung-etal-2020-unsupervised,sun-etal-2021-parallel}.
However, these approaches cannot be used for North Korean alignment as there are no resources immediately available, including, but not limited to, bilingual dictionaries, parallel sentences, and monolingual data. 

\begin{table}
\begin{center}
\begin{tabular}{lccc}
\toprule
  &  dev & test & train \\\midrule
NK-EN & 500 & 500 &  4,109 (4,343) \\
NK-JA & 500 & 500 &  3,739 (3,913) \\\bottomrule
\end{tabular}
\end{center}
\caption{Number of sentences in the manually aligned dev and test data, and the automatically aligned training data. We randomly split 1,000 parallel sentences in half and use them as dev and test data. The numbers in parentheses are those of sentences in the setting that use the bidi-SK for article alignment.}
\label{mt_parallel_corpus}
\end{table}

Although there are some differences between South and North Korean, both forms of Korean have the same basic grammar, and share many vocabularies in common. 
Therefore, a South Korean NMT model can translate North Korean sentences to some extent. 
In this study, inspired by this aspect and a previous approach that used an existing MT model \cite{karimi-etal-2018-extracting}, we design an automatic North Korean alignment method using the South Korean NMT model instead of a North Korean one. 

\paragraph{Sentence alignment.}
\label{sentence_alignment}
In sentence alignment, we assume that parallel North Korean and target language documents are available, defined as $N = \{n_1, ..., n_i\}$ and $T = \{t_1, ..., t_k\}$. Here, $n_i$ corresponds to a sentence in $N$ and $t_k$ for $T$, and $i$, $k$ are the number of sentences in a given document. 

A sentence alignment method based on South Korean NMT consists of two steps. In the first step, we translate $N$ and $T$ into both target language and South Korean using South Korean NMT models. We define the translated documents $N$ and $T$ as $\tilde{N} = \{\tilde{n}_1, ..., \tilde{n}_i\}$  and $\tilde{T} = \{\tilde{t}_1, ..., \tilde{t}_k\}$ and the translated sentences  $n_i$ and $t_k$ as  $\tilde{n}_i$ and $\tilde{t}_k$.  

In the second step, we measure similarity score between the original and translated sentences, and greedily select sentence pairs with the highest similarity score as bilingual sentences. 
An index of bilingual sentence corresponding to $t_k$ is as follows:

\begin{equation}
\hat{j}  =\argmax_{j \in \{1, ..., i\}}\lbrack\mathrm {sim}(\boldsymbol{n_j}, \boldsymbol{\tilde{t}_k}) + \mathrm {sim}(\boldsymbol{\tilde{n}_j}, \boldsymbol{t_k}) \rbrack
\end{equation}
\begin{figure*}[t]
  \begin{minipage}[b]{0.5\hsize}
    \centering
    \includegraphics[scale=0.4]{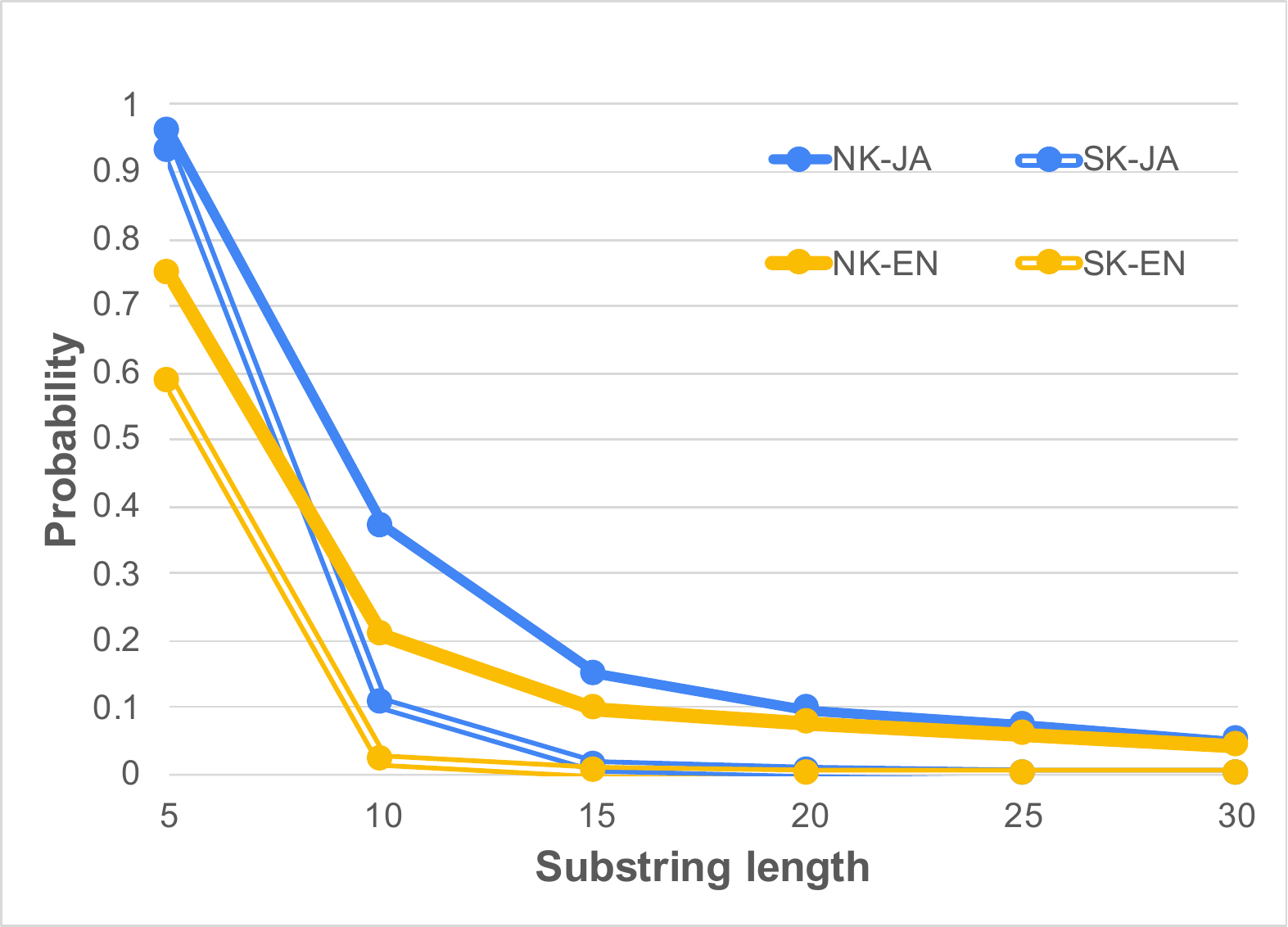}
    \subcaption{in train}\label{dup_train}
  \end{minipage}
  \begin{minipage}[b]{0.5\hsize}
    \centering
    \includegraphics[scale=0.4]{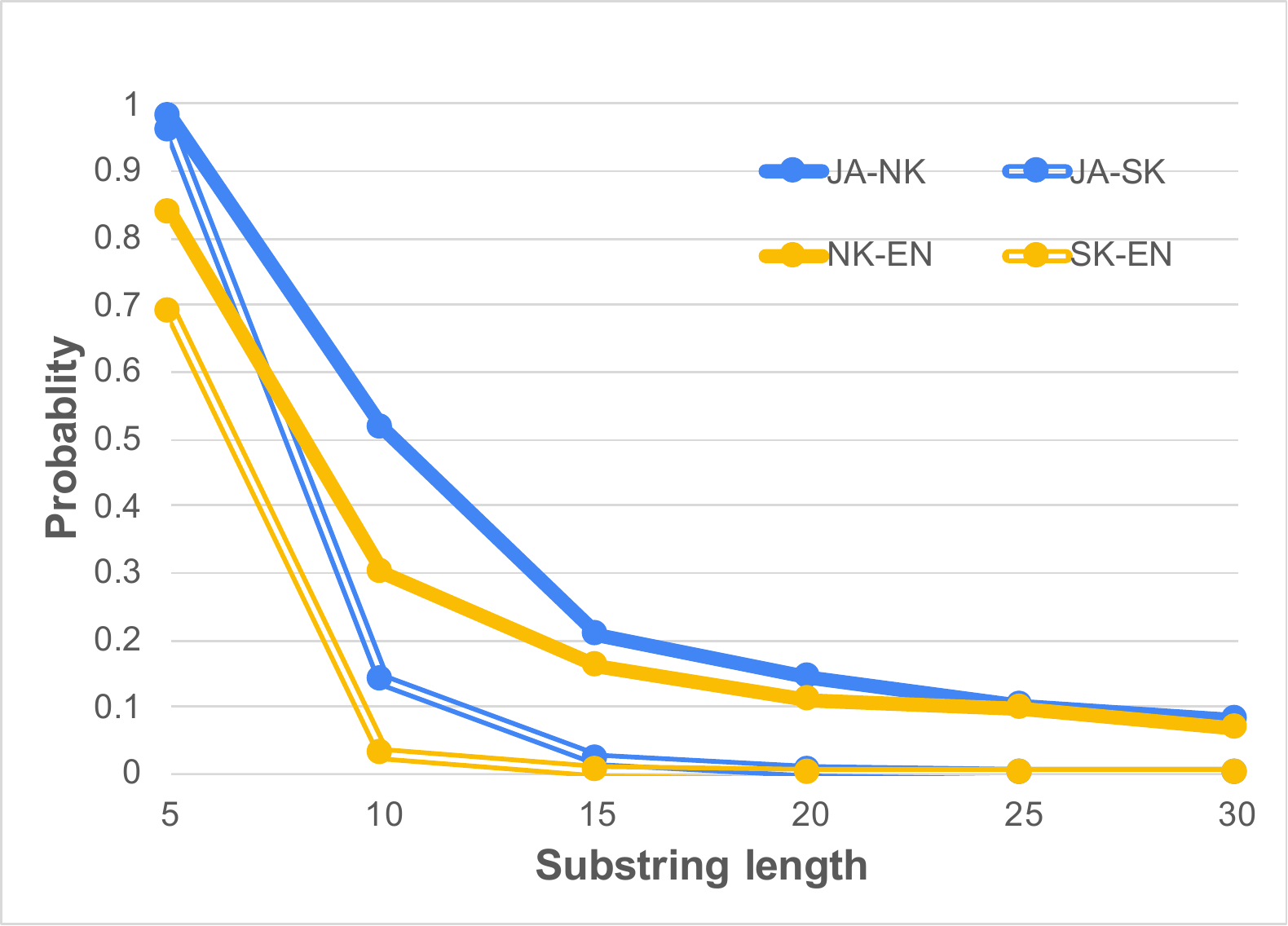}
    \subcaption{between train and dev}\label{dup_dev}
  \end{minipage}
  \caption{Duplication probabilities for each substring length.}\label{dup}
\end{figure*}

\noindent where $\mathrm {sim}$ is a function to measure the similarities between sentence vectors. We use $\mathrm{tf}$-$\mathrm{idf}$  to vectorize a sentence and a margin-based function  \cite{artetxe-schwenk-2019-margin}  as the $\mathrm{sim}$ following LASER \cite{10.1162/tacl_a_00288}. 

In this study, we use bi-directional South Korean NMT models, but this framework can also works in a uni-directional model. We refer to these methods as follows:  to-SK denotes to the South Korean model, from-SK denotes from the South Korean model, and bidi-SK denotes bi-directional South Korean models.

\paragraph{Article alignment.}
\label{sec:article_alignment}
In this study, we also extend this method for aligning articles,
which uses the similarity between the sentences translated by South Korean NMT models. In the article alignment, we use the concatenated sentences of title and document to vectorize each article.

\section{North Korean Alignment Experiments}

\subsection{Experimental Settings}
\label{alignment_experiments}

\paragraph{Manual evaluation data alignment.}

To create the MT evaluation data, we align the articles  (Table \ref{article_statics})  and  1,000 sentences randomly extracted from the English and Japanese articles.
We ask the trilingual annotator to align these articles and sentences.
We also ask the bilingual annotators to align 100 articles and sentences randomly sampled from them. 
Then, we measure the inter-annotator agreement using these 100 articles and sentences.

To create the sentence alignment evaluation data, we chose the article with the most English and Japanese sentences per NK--EN and NK--JA pair. 
The numbers of each language's sentences were  290, 300 and 143, 100 in NK--EN and NK--JA pairs, respectively. 
We ask the trilingual annotator to align these sentences.

\paragraph{Automatic training data alignment.}
We use the news domain translation dataset from AI Hub\footnote{\url{https://aihub.or.kr/}} to train South Korean NMT models. The datasets have 720k and 920k sentences, in SK--EN and SK--JA pairs, respectively.
We pre-tokenize Japanese sentences using MeCab with an IPA dictionary\footnote{\url{https://taku910.github.io/mecab/}}, and then split each language's sentences into subwords using  SentencePiece \citelanguageresource{kudo-richardson-2018-sentencepiece} model with 32k vocabulary size per language. We use a transformer-base \cite{NIPS2017_3f5ee243} for the NMT model using fairseq \citelanguageresource{ott-etal-2019-fairseq}. 

When tokenizing the sentences for the to-SK, from-SK, and bidi-SK,  we train another SentencePiece model using the original sentences and their translations using the South Korean model\footnote{Specifically,  we use sentences of $N$, $\tilde{T}$  and  $\tilde{N}$, $T$  in all articles for each direction, respectively.}. Additionally, we set the vocabulary size to 2k, as there are only a handful of sentences in each article (Table \ref{article_statics}).

We compare the method based on South Korean NMT model to two baselines. 
\begin{inparaenum}[(1)]
\item A naive method aligning the sentences per index in the document. 
\item LASER \cite{10.1162/tacl_a_00288}, which is a cross-lingual model trained by bilingual data between several languages. Notably, North Korean sentences were not included in the training data of LASER.
\end{inparaenum}
When aligning articles using  LASER, we mean-pool each sentence's vectors as LASER to vectorize articles. 

As evaluation metric, we use F1 scores that is an official metric in the BUCC shared task \cite{zweigenbaum-etal-2017-overview}.
Specifically, precision and recall are calculated as percentages of correct pairs among selected and gold pairs.  
We compare each method based on the F1 scores and select the best performing method to align the training data.

\begin{table*}[tp]
\begin{center}
\scalebox{0.95}{
\begin{tabular}{llccccccccc}
\toprule
       & & \multicolumn{2}{c}{SK--EN} & \multicolumn{2}{c}{NK--EN} & \multicolumn{2}{c}{SK--JA} & \multicolumn{2}{c}{NK-JA}   \\ 
   article & model & dev & test & dev & test  & dev & test & dev & test   \\  \midrule
 & SK & \textbf{37.6±.22} &  \textbf{37.7±.20}  & 11.4±.17  & 11.9±.21 & \textbf{71.0±.04}  & \textbf{70.9±.05}  & 36.8±.19 & 37.8±.09 \\  \midrule
   \multirow{3}{*}{Human} & NK & 0.5±.06 & 0.5±.03 & 21.4±.15 & 20.4±.09  & 1.5±.03  & 1.5±.07  & 36.8±.20 & 34.0±.21\\ 
  & SK$\rightarrow$NK  & 11.0±.04 & 11.1±.05 & \textbf{36.7±.12} & \textbf{35.6±.12} & 69.3±.11 & 61.3±.07 & 69.7±.11 & \textbf{69.7±.14} \\  
  & SK$+$NK  & 37.4±.08 & 37.3±.06 & 34.2±.07 & 33.6±.23  & 70.4±.17 & 70.4±.16 & 67.9±.14 & 67.5±.05 \\  \midrule
    \multirow{3}{*}{bidi-SK} & NK & 0.5±.09 & 0.5±.09 & 21.5±.14 & 20.5±.23 & 1.2±.06 & 1.2±.06 & 33.3±.09 & 30.1±.19\\  
  &  SK$\rightarrow$NK   & 24.2±.05 & 24.3±.11 & 36.2±.28 & 35.2±.34  & 47.2±.09 & 47.2±.09 & \textbf{70.5±.14} & 69.4±.26 \\  
  & SK$+$NK & 37.3±.11 & 37.2±.12 & 34.65±.22 & 33.8±.02  & 70.6±.14 & 70.6±.11 & 67.8±.13 & 67.1±.11 \\ 
  \bottomrule
\end{tabular}
}
\end{center}
\caption{
BLEU scores of each model. These BLEU scores are the averages of three models. The rows for Human and bidi-SK are the settings of using human annotation and the bidi-SK for article alignment, respectively. 
}
\label{bleu_score}
\end{table*}

\begin{table*}[tp]
\begin{center}
\scalebox{1.0}{
\begin{tabular}{llcccc}
\toprule
    & & \multicolumn{2}{c}{NK--EN} & \multicolumn{2}{c}{NK--JA}   \\ 
   article & model & dev & test & dev & test  \\  \midrule
& SK  & 9.3±.29  & 9.3±.29 & 39.9±.23 &  39.56±.20  \\  \midrule
     \multirow{3}{*}{Human} & NK & 9.6±.23  & 9.0±.21 & 24.4±.24 & 24.3±.32 \\  
  &  SK$\rightarrow$NK & \textbf{26.0±.15} & \textbf{25.3±.10} & 66.7±.12 & 65.7±.11 \\  
  &  SK$+$NK & 23.5±.16 & 22.5±.34 & 63.4±.15& 63.6±.25   \\  \midrule
    \multirow{3}{*}{bidi-SK} & NK & 8.7±.23 & 7.8±.25 & 22.2±.19 & 22.5±.25\\  
  &  SK$\rightarrow$NK & 25.9±.20 & 24.0±.25 & \textbf{66.9±.12} & \textbf{66.0±.19}  \\  
  &  SK$+$NK & 22.7±.16 & 21.9±.24  & 63.0±.22 & 62.6±.18  \\ 
  \bottomrule
\end{tabular}
}
\end{center}
\caption{
BLEU scores without long substring duplication with the training data.
}
\label{dedup_bleu_score}
\end{table*}

\subsection{Experimental Results}

\paragraph{Manual evaluation data alignment.}
We discuss the results of the article and sentence alignments for the MT evaluation data.
The match rates of the article and sentence alignments between the trilingual and bilingual annotators are 99\%, 95\% and 99\%, 100\% for the NK--EN and NK--JA pairs, respectively.
Based on this, we confirm that the aligned articles and sentences are in agreement between the annotators. 
As the results of manual alignments, we obtain 359 and 356 parallel articles and 1,000 parallel sentences  for the NK--EN and NK--JA pairs, respectively.
We also obtain evaluation data for automatic sentence alignment that consisted of  285 and 100 parallel sentences through the alignments of the sentences of a parallel article chosen per NK--EN and NK--JA pairs.
We summarize the evaluation data for sentence and article alignments in Table \ref{manual_alignment}.

\paragraph{Automatic training data alignment.}
We show the alignment quality of each methods using the manually created evaluation data (Table \ref{manual_alignment}).
Table \ref{sentence_alignment_acc2} shows the F1 scores of each sentence alignment method.
LASER, a strong baseline, achieves  96.8  and 96.9 in each language pair, respectively, and significantly outperforms the naive method. 
The to-SK and from-SK also align the sentences with high scores. The bidi-SK further improves the F1 scores and slightly outperforms the LASER. 
Table \ref{sentence_alignment_acc2} also shows the article alignment F1 scores of each method. The bidi-SK achieves the highest scores of 97.6  and 98.4 for the NK--EN and NK--JA pairs. 
These results show that bidi-SK is suitable for North Korean sentence and article alignment.

Therefore, we adopt the bidi-SK for aligning North Korean training data. We exclude the sentences included in the manually created evaluation data from the documents, and then, apply the bidi-SK to align parallel sentences.
A summary of our North Korean parallel corpus constructed through manual and automatic alignment is presented in Table \ref{mt_parallel_corpus}. 

\paragraph{Characteristic of North Korean parallel corpus.}
\label{characteristic}
We discuss the characteristics of the North Korean parallel corpus.
Owing  to the nature of North Korean articles, our corpus contains a significant amount of duplicated substrings. Following Lee et al. \cite{lee2021}, we measure the duplication probability of word substrings in South and North Korean training data of English and Japanese sides. 
The duplication probabilities for each substring length are in Figure \ref{dup_train}. This figure indicates that the probabilities of substrings with more than ten consecutive duplicate words are higher in North Korean than those of South Korean data. 
We also calculate the duplication probabilities between the dev and train sentences as shown Figure \ref{dup_dev}. It indicates that North Korean evaluation data contains many duplicates of long substrings with training data. 

Thus, our North Korean parallel corpus has some limitations regarding the size and diversity of sentences. However, our corpus is useful for developing a North Korean translation system because there is no other corpus with such data available. Additionally, the results of the automatic alignment experiments will serve as a useful reference when creating more parallel corpora in the future.

\begin{table*}
\begin{center}
\scalebox{1.0}{
\begin{tabular}{ll}
\toprule
\small
NK source & $\ldots$ 제7기 제6차\textbf{전원회의} 결정을 관철하는데서 $\ldots$  \\
Reference & $\ldots$ implementing the decision of the 6th \textbf{Plenary Meeting} of the 7th  $\ldots$
 \\\midrule
SK  &  $\ldots$  carrying out the decision of the 7th full session of $\ldots$  \\
NK & $\ldots$  decided on the issue of convening the 4th \textbf{Plenary Meeting} of the 8th  $\ldots$  \\
SK$\rightarrow$NK & $\ldots$  implementing the decision of the Sixth \textbf{Plenary Meeting} of the Seventh $\ldots$  \\
 SK$+$NK & $\ldots$ implementing the decision of the 6th \textbf{Plenary Meeting} of the 7th $\ldots$\\
Google & $\ldots$  carrying out the decision of the 7th 6th \textbf{Plenary Meeting} of  $\ldots$  \\
Naver & $\ldots$  carrying out the decision of the 7th 6th session of $\ldots$ \\
\bottomrule
\end{tabular}
}
\end{center}
\caption{
Translation example.
}
\label{translation2}
\end{table*}

\section{North Korean NMT Experiments}

\subsection{Experimental Settings}
We use the same South Korean datasets and implementation of the NMT model as presented in Section \ref{alignment_experiments}.
We use merged sentences of South and North Korean bilingual data for training the SentencePiece model, and set the vocabulary size as 32k per language.

We compare the models trained by only South or North Korean data (SK and NK),  a South Korean model fine-tuned by North Korean data (SK$\rightarrow$NK), and a combined model jointly trained by South and North Korean data (SK$+$NK). 
We also investigate translation accuracy of the models that use the bilingual articles aligned by our proposed method. 

\subsection{Experimental Results}

\paragraph{Quantitative evaluation.}
Table \ref{bleu_score} shows the BLEU scores of each model. The SK$\rightarrow$NK model achieves the highest BLEU scores in the evaluations of  NK--EN and NK--JA pairs. This result indicates that the models trained by only SK or NK data cannot translate North Korean sentences well, but fine-tuning the SK model using a small amount of North Korean data can significantly boost the translation quality. 
The SK$+$NK model also improves the BLEU scores 
and mitigates degradation in the SK--EN and SK--JA evaluations compared to the SK$\rightarrow$NK model. 
Therefore, we consider that improving the  SK$+$NK model is a good way to develop a universal Korean NMT model.
In addition, surprisingly, the models that use our method for aligning the articles achieve similar scores as the models that use human annotation. 

As discussed in Subsection \ref{characteristic}, the North Korean evaluation data contains many duplicates of long substrings with training data. Because duplicates of long substrings between the train and evaluation data leads to an overestimation of the North Korean models, we evaluate the models by deleting the sentences of dev and test that duplicate more than ten substrings with the train data.
Table \ref{dedup_bleu_score} shows the BLEU scores obtained using the North Korean evaluation data without long substring duplication. The BLEU scores of the SK models are almost the same, whereas those of the NK, SK$\rightarrow$NK, and  SK$+$NK models have decreased by 4--10 points compared to the scores in Table \ref{bleu_score}. However, the relations between the BLEU scores of each model show the same trend. 

\paragraph{Qualitative evaluation.} 
The most significant advantage of using the North Korean data is the ability to translate words that are spelled differently or have semantically diverged in South Korea. 
For example, the word \textit{Vladivostok} is written as ``블라디보스토크'' in South Korea but ``울라지보스또크'' in North Korea. Therefore, whereas publicly available South Korean models such as those of Google\footnote{\url{https://translate.google.com}} and NAVER\footnote{\url{https://papago.naver.com}} are unable to translate  ``울라지보스또크,'' which means \textit{Vladivostok}, the models trained with North Korean data are able to produce correct translations (Table \ref{translation}). On the other hand, the compound word \textit{plenary meeting} is written as ``전원회의'' in North Korean. Both the words ``전원'' and ``회의'' are also used in South Korean, but these words are not used in conjunction as \textit{plenary meeting}.
The South Korean models translate ``전원회의'' to \textit{full session}, which is similar, but not perfect. In contrast, the models using North Korean data translate it correctly (Table  \ref{translation2}). Notably, the NK model, which uses a small amount of North Korean data, translates ``전원회의'' appropriately, but it cannot translate fluently. Specifically, the NK model translates ``제7기'' and ``제6차'' to \textit{4th} and \textit{8th}, respectively, and repeats the words \textit{decided on the issue of convening} (Table \ref{translation2}).

\section{Conclusion}
In this study, we manually created evaluation data for automatic alignment and MT systems. Moreover, we showed that bidi-SK is suitable for the alignment of North Korean parallel sentences, and constructed North Korean training data using bidi-SK.
Finally, we demonstrated that our training data can enhance North Korean translation quality. 
Although our North Korean MT datasets  have some limitations with regards to size and diversity of sentences, the findings of our study are useful for the development of a North Korean translation system. To support further research, we also provide the data and code used in our experiments.

Our translation experiments also indicated a trade-off between the accuracy of South and North Korean translations. Therefore, a universal Korean NMT system that can handle both Korean language variants is still an open problem to be solved.

\section{Bibliographical References}\label{reference}

\bibliographystyle{lrec2022-bib}
\bibliography{lrec2022-example}

\begin{thebibliography}{}

\bibitem[\protect\citename{Artetxe and
  Schwenk}2019a]{artetxe-schwenk-2019-margin}
Artetxe, M. and Schwenk, H.
\newblock (2019a).
\newblock Margin-based parallel corpus mining with multilingual sentence
  embeddings.
\newblock In {\em Proceedings of the 57th Annual Meeting of the Association for
  Computational Linguistics}.

\bibitem[\protect\citename{Artetxe and Schwenk}2019b]{10.1162/tacl_a_00288}
Artetxe, M. and Schwenk, H.
\newblock (2019b).
\newblock Massively multilingual sentence embeddings for zero-shot
  cross-lingual transfer and beyond.
\newblock {\em Transactions of the Association for Computational Linguistics},
  7:597--610, 09.

\bibitem[\protect\citename{Azpeitia \bgroup et al.\egroup
  }2017]{azpeitia-etal-2017-weighted}
Azpeitia, A., Etchegoyhen, T., and Mart{\'\i}nez~Garcia, E.
\newblock (2017).
\newblock Weighted set-theoretic alignment of comparable sentences.
\newblock In {\em Proceedings of the 10th Workshop on Building and Using
  Comparable Corpora}.

\bibitem[\protect\citename{Chen}1993]{chen-1993-aligning}
Chen, S.~F.
\newblock (1993).
\newblock Aligning sentences in bilingual corpora using lexical information.
\newblock In {\em Proceedings of 31st Annual Meeting of the Association for
  Computational Linguistics}.

\bibitem[\protect\citename{Dabre \bgroup et al.\egroup
  }2017]{dabre-etal-2017-empirical}
Dabre, R., Nakagawa, T., and Kazawa, H.
\newblock (2017).
\newblock An empirical study of language relatedness for transfer learning in
  neural machine translation.
\newblock In {\em Proceedings of the 31st Pacific Asia Conference on Language,
  Information and Computation}.

\bibitem[\protect\citename{Durrani \bgroup et al.\egroup
  }2010]{durrani-etal-2010-hindi}
Durrani, N., Sajjad, H., Fraser, A., and Schmid, H.
\newblock (2010).
\newblock {H}indi-to-{U}rdu machine translation through transliteration.
\newblock In {\em Proceedings of the 48th Annual Meeting of the Association for
  Computational Linguistics}.

\bibitem[\protect\citename{Etchegoyhen and
  Azpeitia}2016]{etchegoyhen-azpeitia-2016-set}
Etchegoyhen, T. and Azpeitia, A.
\newblock (2016).
\newblock Set-theoretic alignment for comparable corpora.
\newblock In {\em Proceedings of the 54th Annual Meeting of the Association for
  Computational Linguistics (Volume 1: Long Papers)}.

\bibitem[\protect\citename{Gomes and Lopes}2016]{gomes-lopes-2016-first}
Gomes, L. and Lopes, G.~P.
\newblock (2016).
\newblock First steps towards coverage-based sentence alignment.
\newblock In {\em Proceedings of the Tenth International Conference on Language
  Resources and Evaluation ({LREC}'16)}.

\bibitem[\protect\citename{Harrat \bgroup et al.\egroup }2019]{HARRAT2019262}
Harrat, S., Meftouh, K., and Smaili, K.
\newblock (2019).
\newblock Machine translation for arabic dialects (survey).
\newblock {\em Information Processing \& Management}, 56(2):262--273.

\bibitem[\protect\citename{Karimi \bgroup et al.\egroup
  }2018]{karimi-etal-2018-extracting}
Karimi, A., Ansari, E., and Sadeghi~Bigham, B.
\newblock (2018).
\newblock Extracting an {E}nglish-{P}ersian parallel corpus from comparable
  corpora.
\newblock In {\em Proceedings of the Eleventh International Conference on
  Language Resources and Evaluation ({LREC} 2018)}.

\bibitem[\protect\citename{Keung \bgroup et al.\egroup
  }2020]{keung-etal-2020-unsupervised}
Keung, P., Salazar, J., Lu, Y., and Smith, N.~A.
\newblock (2020).
\newblock Unsupervised bitext mining and translation via self-trained
  contextual embeddings.
\newblock {\em Transactions of the Association for Computational Linguistics},
  8:828--841.

\bibitem[\protect\citename{Kim \bgroup et al.\egroup }2020]{kim-etal-2020-zero}
Kim, H., Hirasawa, T., and Komachi, M.
\newblock (2020).
\newblock Zero-shot {N}orth {K}orean to {E}nglish neural machine translation by
  character tokenization and phoneme decomposition.
\newblock In {\em Proceedings of the 58th Annual Meeting of the Association for
  Computational Linguistics: Student Research Workshop}.

\bibitem[\protect\citename{Lakew \bgroup et al.\egroup
  }2018]{lakew-etal-2018-neural}
Lakew, S.~M., Erofeeva, A., and Federico, M.
\newblock (2018).
\newblock Neural machine translation into language varieties.
\newblock In {\em Proceedings of the Third Conference on Machine Translation:
  Research Papers}.

\bibitem[\protect\citename{Lee \bgroup et al.\egroup }2021]{lee2021}
Lee, K., Ippolito, D., Nystrom, A., Zhang, C., Eck, D., Callison-Burch, C., and
  Carlini, N.
\newblock (2021).
\newblock Deduplicating training data makes language models better.
\newblock {\em arXiv preprint arXiv:2107.06499}.

\bibitem[\protect\citename{Lee}1990]{Lee1990}
Lee, H.~B.
\newblock (1990).
\newblock {Differences in language use between North and South Korea}.
\newblock {\em International Journal of the Sociology of Language},
  1990(82):71--86.

\bibitem[\protect\citename{Marujo \bgroup et al.\egroup
  }2011]{marujo-etal-2011-bp2ep}
Marujo, L., Grazina, N., Luis, T., Ling, W., Coheur, L., and Trancoso, I.
\newblock (2011).
\newblock {BP}2{EP} - adaptation of {B}razilian {P}ortuguese texts to
  {E}uropean {P}ortuguese.
\newblock In {\em Proceedings of the 15th Annual conference of the European
  Association for Machine Translation}.

\bibitem[\protect\citename{Nakazawa \bgroup et al.\egroup
  }2021]{nakazawa-etal-2021-overview}
Nakazawa, T., Nakayama, H., Ding, C., Dabre, R., Higashiyama, S., Mino, H.,
  Goto, I., Pa, W.~P., Kunchukuttan, A., Parida, S., Bojar, O., Chu, C.,
  Eriguchi, A., Abe, K., and Oda, Yusuke~Kurohashi, S.
\newblock (2021).
\newblock Overview of the 8th workshop on {A}sian translation.
\newblock In {\em Proceedings of the 8th Workshop on Asian Translation}.

\bibitem[\protect\citename{Popovi{\'c} \bgroup et al.\egroup
  }2016]{popovic-etal-2016-language}
Popovi{\'c}, M., Ar{\v{c}}an, M., and Klubi{\v{c}}ka, F.
\newblock (2016).
\newblock Language related issues for machine translation between closely
  related {S}outh {S}lavic languages.
\newblock In {\em Proceedings of the Third Workshop on {NLP} for Similar
  Languages, Varieties and Dialects ({V}ar{D}ial3)}.

\bibitem[\protect\citename{Schwenk and Douze}2017]{schwenk-douze-2017-learning}
Schwenk, H. and Douze, M.
\newblock (2017).
\newblock Learning joint multilingual sentence representations with neural
  machine translation.
\newblock In {\em Proceedings of the 2nd Workshop on Representation Learning
  for {NLP}}.

\bibitem[\protect\citename{Schwenk}2018]{schwenk-2018-filtering}
Schwenk, H.
\newblock (2018).
\newblock Filtering and mining parallel data in a joint multilingual space.
\newblock In {\em Proceedings of the 56th Annual Meeting of the Association for
  Computational Linguistics (Volume 2: Short Papers)}.

\bibitem[\protect\citename{Sennrich and Volk}2010]{sennrich-volk-2010-mt}
Sennrich, R. and Volk, M.
\newblock (2010).
\newblock {MT}-based sentence alignment for {OCR}-generated parallel texts.
\newblock In {\em Proceedings of the 9th Conference of the Association for
  Machine Translation in the Americas: Research Papers}.

\bibitem[\protect\citename{Sun \bgroup et al.\egroup
  }2021]{sun-etal-2021-parallel}
Sun, Y., Zhu, S., Yifan, F., and Mi, C.
\newblock (2021).
\newblock Parallel sentences mining with transfer learning in an unsupervised
  setting.
\newblock In {\em Proceedings of the 2021 Conference of the North American
  Chapter of the Association for Computational Linguistics: Student Research
  Workshop}.

\bibitem[\protect\citename{Tan \bgroup et al.\egroup }2012]{6473708}
Tan, T.-P., Goh, S.-S., and Khaw, Y.-M.
\newblock (2012).
\newblock A {Malay} dialect translation and synthesis system: Proposal and
  preliminary system.
\newblock In {\em 2012 International Conference on Asian Language Processing}.

\bibitem[\protect\citename{Vaswani \bgroup et al.\egroup
  }2017]{NIPS2017_3f5ee243}
Vaswani, A., Shazeer, N., Parmar, N., Uszkoreit, J., Jones, L., Gomez, A.~N.,
  Kaiser, L.~u., and Polosukhin, I.
\newblock (2017).
\newblock Attention is all you need.
\newblock In {\em Advances in Neural Information Processing Systems}.

\bibitem[\protect\citename{Yun and Kang}2019]{YUN2019100918}
Yun, S. and Kang, Y.
\newblock (2019).
\newblock Variation of the word-initial liquid in {North and South Korean}
  dialects under contact.
\newblock {\em Journal of Phonetics}, 77:100918.

\bibitem[\protect\citename{Zoph \bgroup et al.\egroup
  }2016]{zoph-etal-2016-transfer}
Zoph, B., Yuret, D., May, J., and Knight, K.
\newblock (2016).
\newblock Transfer learning for low-resource neural machine translation.
\newblock In {\em Proceedings of the 2016 Conference on Empirical Methods in
  Natural Language Processing}.

\bibitem[\protect\citename{Zweigenbaum \bgroup et al.\egroup
  }2017]{zweigenbaum-etal-2017-overview}
Zweigenbaum, P., Sharoff, S., and Rapp, R.
\newblock (2017).
\newblock Overview of the second {BUCC} shared task: Spotting parallel
  sentences in comparable corpora.
\newblock In {\em Proceedings of the 10th Workshop on Building and Using
  Comparable Corpora}.

\end{thebibliography}

\section{Language Resource References}
\label{lr:ref}
\bibliographystylelanguageresource{lrec2022-bib}
\bibliographylanguageresource{languageresource}

\end{document}